\documentclass{nature}
\usepackage{graphicx}
\graphicspath{ {./img/} }
\usepackage{lineno}
\usepackage{booktabs}
\usepackage{float}
\usepackage{ulem}
\usepackage{gensymb}

\usepackage{hyphenat}
\usepackage{xcolor}
\usepackage{bm}
\usepackage{hyperref}
\hypersetup{
    colorlinks = true,
    linkbordercolor = {white},
    citecolor = black,
    urlcolor  = black,
    linkcolor = black,
}
\usepackage{amsmath}  
\usepackage{amsfonts}
\usepackage{subfigure}
\usepackage{float}
\usepackage{caption}
\makeatletter
\let\saved@includegraphics\includegraphics
\AtBeginDocument{\let\includegraphics\saved@includegraphics}
\renewenvironment*{figure}{\@float{figure}}{\end@float}
\makeatother

\title{
\parbox{\textwidth}
{\centering From Prediction to Diagnosis: Reasoning-Aware AI for Photovoltaic Defect Inspection} 
}

\author{Dev Mistry$^{1}$, Feng Qiu$^{2}$, Bo Chen$^{3}$, Feng Liu$^{4}$, Can Chen$^{5}$, Mohammad Shahidehpour$^{1}$, Ren Wang$^{1,*}$}

\begin{document}

\maketitle 

\begin{affiliations}
\footnotesize

 \item Department of Electrical and Computer Engineering, Illinois Institute of Technology, Chicago, IL 60616, USA
 \item Argonne National Laboratory, Lemont, IL 60439, USA
 \item Smart Grid Emerging Technology Department, Commonwealth Edison, Chicago, IL 60618, USA
 \item Department of Computer Science, Drexel University, Philadelphia, PA 19104, USA
 \item School of Data Science and Society, University of North Carolina at Chapel Hill, Chapel Hill, NC 27599, USA
 \item[* ] To whom correspondence should be addressed (rwang74@illinoistech.edu).
\end{affiliations}

\newpage

\section{Abstract}
Reliable photovoltaic defect identification is essential for maintaining energy yield, ensuring warranty compliance, and enabling scalable inspection of rapidly expanding solar fleets. Although recent advances in computer vision have improved automated defect detection, most existing systems operate as opaque classifiers that provide limited diagnostic insight for high-stakes energy infrastructure. Here we introduce REVL-PV, a vision-language framework that embeds domain-specific diagnostic reasoning into multimodal learning across electroluminescence, thermal, and visible-light imagery. By requiring the model to link visual evidence to plausible defect mechanisms before classification, the framework produces structured diagnostic reports aligned with professional photovoltaic inspection practice. Evaluated on 1,927 real-world modules spanning eight defect categories, REVL-PV achieves 93\% classification accuracy while producing interpretable diagnostic rationales and maintaining strong robustness under realistic image corruptions. A blind concordance study with a certified solar inspection expert shows strong semantic alignment between model explanations and expert assessments across defect identification, root-cause attribution, and visual descriptions. These results demonstrate that reasoning-aware multimodal learning establishes a general paradigm for trustworthy AI-assisted inspection of photovoltaic energy infrastructure.

\newpage

The global deployment of solar photovoltaics is accelerating at an exponential pace. Cumulative installed PV capacity worldwide surpassed 2.26 TW by the end of 2024 after record annual additions of 553-601 GW, more than any other energy-generation technology\cite{iea-pvps-trends2025}. According to the International Energy Agency, renewable power capacity is set to expand by nearly 4,600 GW between 2025 and 2030 (equivalent to the combined current power capacity of China, the European Union and Japan), with solar PV accounting for almost 80 percent of this growth\cite{iea-renewables2025}. This scale implies that hundreds of millions of new modules will enter operation annually worldwide. Thus, with this unprecedented expansion comes the critical challenge that undetected or misclassified defects will accelerate degradation, reduce energy yields across utility-scale fleets by several percent, increase operations-and-maintenance costs, and create safety hazards including fires and electrical faults. As a result, the exclusive reliance on manual inspection has become increasingly untenable, intensifying the need for automated diagnostic systems that can operate reliably across heterogeneous imaging conditions and defect types while effectively supporting and augmenting human expertise. 

Earlier methods based on thresholding with morphological operations\cite{hu2024research}, Fourier image reconstruction\cite{tsai2012defect} and handcrafted reconstruction pipelines were effective only for obvious anomalies under controlled conditions, but they lacked robustness to illumination changes, subtle internal faults, and large-scale deployment variability\cite{hijjawi2023review}. More recent Convolutional Neural Network- and Vision Transformer (ViT)-based architectures have substantially improved defect localization and classification performance. Among all these methods, the YOLO family of models\cite{mohamed2025optimized, al2025effective, li2025yolo} and various ViT frameworks\cite{rodrigo2025vision, khan2025deep} have become particularly prominent for module-level fault detection. However, most existing approaches still function primarily as black-box predictors. They exhibit reduced generalization under varying field imaging conditions, provide limited interpretability and little support for structured diagnostic decision-making. Even early multimodal frameworks, such as the PVAL\cite{guo2025solar} system, is optimized for satellite-level array localization rather than detailed module-level diagnostic reasoning. In energy applications where inspection outcomes inform maintenance actions, warranty claims, and operational risk management, such opacity poses a significant barrier to adoption.

On the other hand, human inspectors do not diagnose PV defects through single-step classification. Instead, they follow structured diagnostic reasoning that links observable visual cues to plausible physical mechanisms before determining defect type and recommended action. This intermediate reasoning process is critical for resolving ambiguity, auditing decisions, and maintaining trust in inspection outcomes. We hypothesize that the gaps in current automated inspection systems stem from a mismatch between how models are trained and how photovoltaic defects are diagnosed in practice. Bridging this gap requires AI systems that ground multimodal learning in expert domain knowledge, treating diagnostic reasoning as an integral prerequisite for classification rather than a post-hoc explanation.

To address these challenges, we introduce a Reasoning-Boosted Vision-Language for Photovoltaics (REVL-PV) framework (Fig.~\ref{fig:model_architecture}) that integrates four tightly coupled components. First, we apply class-balanced data curation together with physically motivated data augmentation to mitigate long-tailed defect distributions and imaging variability common in photovoltaic manufacturing and field inspection. Second, the system employs structured diagnostic prompting that explicitly separates intermediate reasoning from final defect classification. Third, model adaptation is performed in a staged manner, first establishing domain-consistent diagnostic logic and subsequently grounding this logic in visual evidence across heterogeneous imaging modalities, including EL, thermal, and RGB imagery, each of which encodes distinct physical information about module health. Finally, a reasoning-stabilized inference strategy using test-time augmentation (TTA). The resulting diagnostic outputs extend beyond defect labels to include visual evidence description, inferred root cause, and recommended inspection actions, aligning automated predictions with professional photovoltaic inspection workflows.

A central design objective of this work is practical deployability within energy infrastructure. Rather than relying on large, general-purpose foundation models, we demonstrate that compact vision-language models (VLM) equipped with domain-specific reasoning can achieve high diagnostic accuracy while remaining compatible with the memory and reliability constraints of industrial inspection systems. This balance is essential for real-world photovoltaic inspection, where decisions must be made consistently and with clear justification. By integrating reasoning directly into learning rather than adding it post hoc, REVL-PV improves defect classification, particularly for rare and ambiguous categories. Evaluated on 1,927 real-world photovoltaic modules, the framework achieves 93\% accuracy while generating transparent diagnostic rationales. These results show that compact, domain-adapted vision-language models can provide a practical alternative to larger general-purpose systems for photovoltaic inspection, supporting both autonomous screening and human-in-the-loop workflows. A blind comparison against a certified solar panel expert further shows that the model's reasoning and final diagnoses align closely with professional standards, highlighting the broader potential of reasoning-aware compact VLMs for industrial diagnostics.

\section*{Main Text}

\begin{figure}[H]
    \phantomsection
    \captionsetup{
        justification=raggedright,
        singlelinecheck=false,
        labelfont=bf
    }

    \makebox[\textwidth][c]{%
        \includegraphics[width=1\textwidth]{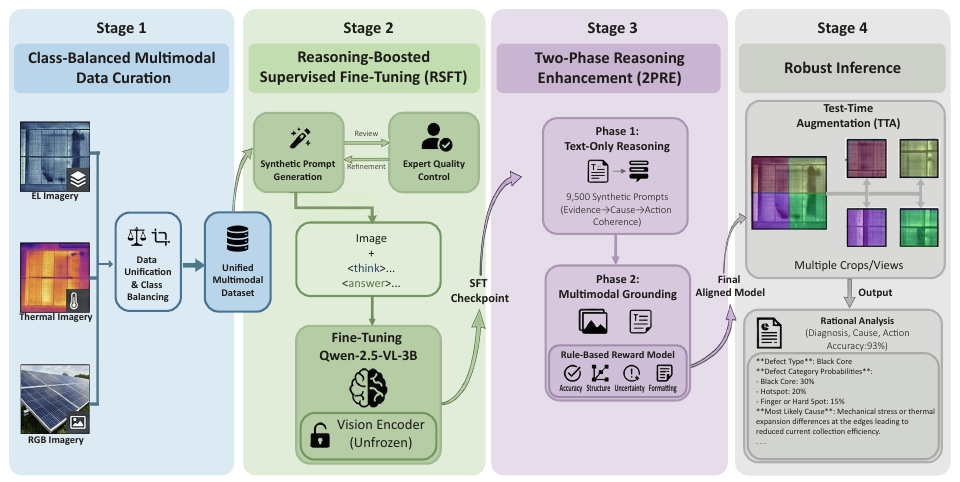}
    }

    \caption{\textbf{REVL-PV model overview.} 
\textbf{Stage 1: Class-Balanced Multimodal Data Curation.} Photovoltaic inspection data are inherently heterogeneous, with defect categories differing in visual distinctiveness and imaging modalities (EL, thermal, visible-light) varying in diagnostic quality. To address this, inspection images from multiple public datasets are unified, preprocessed, and curated using class balancing and physically motivated augmentation to ensure consistent representation of rare and subtle defects across modalities.
\textbf{Stage 2: Reasoning-Boosted Supervised Fine-Tuning (RSFT).} The vision-language backbone is fine-tuned using generated, reasoning-dense samples to embed diagnostic logic via \texttt{<think>} and \texttt{<answer>} tokens.
\textbf{Stage 3: Two-Phase Reasoning Enhancement (2PRE).} We establish text-only \textit{Evidence} $\rightarrow$ \textit{Cause} $\rightarrow$ \textit{Action} coherence, then ground this logic visually using rule-based reinforcement learning.
\textbf{Stage 4: Robust Inference.} Test-Time Augmentation (TTA) aggregates multi-crop predictions to finalize the defect diagnosis and its underlying rationale.}

    \label{fig:model_architecture}
\end{figure}

\noindent{}\textbf{A brief overview of REVL-PV.} We present REVL-PV, a domain-aware, reasoning-based vision-language framework for photovoltaic defect diagnosis (Fig.~\ref{fig:model_architecture}). Rather than treating defect classification as direct pattern recognition, REVL-PV integrates photovoltaic-specific domain knowledge with structured diagnostic reasoning, mirroring the Evidence→Cause→Action logic employed by human inspectors. The framework introduces two key components: (1) Reasoning-Boosted Supervised Fine-Tuning (RSFT), which explicitly conditions classification on intermediate diagnostic steps linking visual observations to fault mechanisms (Fig.~\ref{fig:model_architecture} Stage 2) and (2) a Two-Phase Reasoning Enhancement strategy (2PRE), which first aligns reasoning structure through text-only training before grounding it to visual features via multimodal grounding (Fig.~\ref{fig:model_architecture} Stage 3). Together, these components enable transparent, auditable defect diagnosis without sacrificing classification accuracy.

At its core, the REVL-PV framework is built upon a multimodal vision-language architecture designed to capture the complex, multi-physics nature of photovoltaic failures. It synthesizes complementary physical information by integrating EL for electrical recombination faults, thermal for hotspot anomalies, and RGB imagery for visual surface degradation. Given that real-world inspection data is inherently skewed, we first curate a class-balanced multimodal dataset (Fig.~\ref{fig:model_architecture} Stage 1). By employing quota-based sampling and augmentation, we ensure that rare defects achieve representation parity with common categories, compelling the model to maintain diagnostic sensitivity across the full spectrum of severities rather than optimizing solely for frequent failure modes. Building on this foundation, the curated images from this unified dataset are first utilized to generate synthetic reasoning prompts. These prompts are subsequently subjected to rigorous quality control. In particular, a random sample of 500 generated prompts was manually reviewed by an expert to identify potential hallucinations and false reasoning traces, as well as to verify their consistency with known defect mechanisms and standard inspection practices. This verification step ensures that the reasoning supervision reflects realistic diagnostic logic rather than purely synthetic explanations. Following this validation of the generation process, the complete set of image-prompt pairs is then used to initialize the model through RSFT (Fig.~\ref{fig:model_architecture} Stage 2). This domain-aware, reasoning-based supervision explicitly separates the diagnostic process from final defect classification. Rather than producing a defect label directly, the framework requires the model to articulate intermediate steps that identify salient visual evidence, connect it to plausible fault mechanisms, and justify the resulting diagnosis. By doing so, the framework encourages decisions that are traceable and auditable, mirroring professional inspection practice. Following initialization, REVL-PV refines diagnostic outputs using the 2PRE (Fig.~\ref{fig:model_architecture} Stage 3) approach. This module first performs text-only reasoning on synthetic diagnostic prompts, then applies multimodal grounding to anchor these reasoning chains to specific visual features, allowing the framework to generate diagnostic rationales that are both logically structured and visually grounded. Finally, robust inference is achieved by leveraging TTA (Fig.~\ref{fig:model_architecture} Stage 4) with multi-view prediction aggregation, enabling consistent inspection outcomes under variations in viewpoint, illumination, and acquisition conditions commonly encountered in manufacturing lines and field deployments. Together, these elements embed diagnostic reasoning directly into the inspection process, yielding accurate and transparent photovoltaic defect diagnoses aligned with operational energy-system requirements.

\begin{figure}[H]
    \phantomsection
    \captionsetup{
        justification=raggedright,
        singlelinecheck=false,
        labelfont=bf
    }

    \makebox[\textwidth][c]{%
        \includegraphics[width=0.9\textwidth]{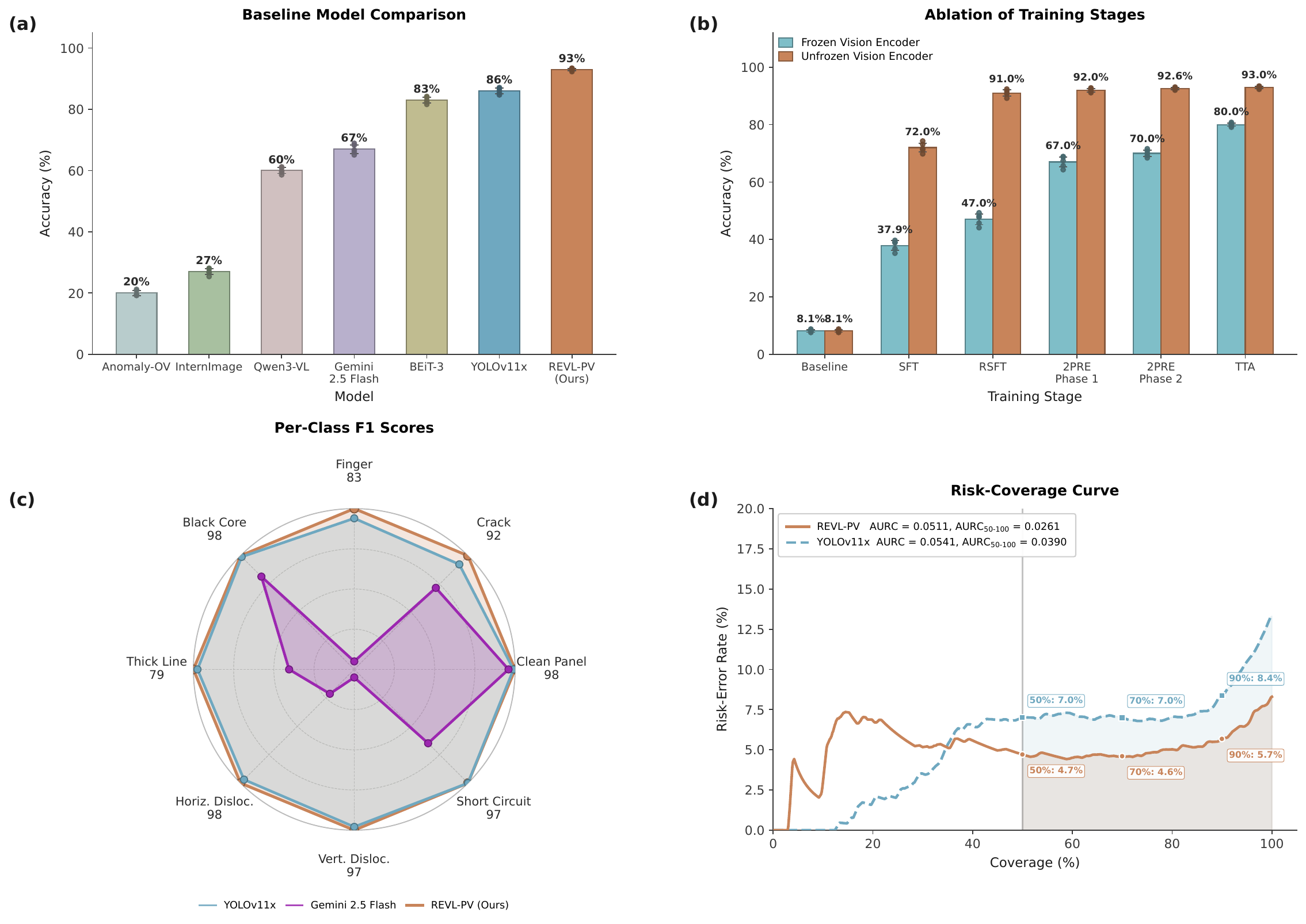}
    }
    
    \caption{\textbf{Quantitative evaluation of the REVL-PV framework.}
\textbf{a. Baseline Model Comparison.} The proposed method achieves a peak accuracy of 93\%, outperforming both general vision models and specialized object detectors like YOLOv11x.
\textbf{b. Ablation of Training Stages.} SStep-by-step performance gains illustrate the compounding value of the reasoning pipeline. While standard supervised fine-tuning (SFT) reaches only 72.0\%, the transition to RSFT and subsequent 2PRE phases systematically refines the model's reasoning capabilities, steadily driving accuracy to 93.0
\textbf{c. Per-Class F1 Scores.} REVL-PV exhibits a consistently superior performance envelope across all defect categories.
\textbf{d. Risk-Coverage Curve.} In the operationally critical high-coverage regime ($\ge$50\%, shaded) most relevant to automated inspection deployment, REVL-PV avoids the escalating overconfidence of YOLOv11x, maintaining a highly stable risk profile and a superior partial AURC\textsubscript{50-100}.}

    \label{fig:model_accuracy_comparison}
\end{figure}

\noindent{}\textbf{REVL-PV outperforms existing methods in defect classification.} To evaluate the performance of the REVL-PV framework, we assessed classification accuracy on a held-out test set of 1,927 images spanning eight defect categories (clean panel, crack, short circuit, thick line, horizontal dislocation, vertical dislocation, finger, and black core). We benchmarked the framework against six representative baselines, all of which were explicitly fine-tuned on our dataset. These baselines include YOLOv11x\cite{khanam2024yolov11}, BEiT-3\cite{wang2023image}, InternImage\cite{wang2023internimage}, Anomaly-OV\cite{xu2025towards}, Gemini 2.5 Flash\cite{comanici2025gemini}, Qwen3-VL\cite{bai2025qwen3}. These models span object detection, vision transformers, convolutional architectures, vision-language anomaly detection, and general multimodal large language model paradigms (Fig.~\ref{fig:model_accuracy_comparison}a). REVL-PV achieves higher classification accuracy than representative baselines, reaching an overall accuracy of 93\%. Compared with the strongest baseline, YOLOv11x\cite{khanam2024yolov11} (86\%), this represents a 7 percentage point improvement, demonstrating the benefits of incorporating domain-aware diagnostic reasoning into defect classification. The performance margins are even more pronounced over BEiT-3\cite{wang2023image} (83\%), Gemini 2.5 Flash\cite{comanici2025gemini} (67\%), Qwen-3-VL\cite{bai2025qwen3} (60\%), InternImage\cite{wang2023internimage} (27\%), and Anomaly-OV\cite{xu2025towards} (20\%). Extended descriptions of all baseline architectures are provided in Supplementary Note 1.

\noindent{}\textbf{REVL-PV establishes structured reasoning as the definitive driver of performance.} To isolate the specific contribution of reasoning-aware supervision, we conducted ablation experiments comparing REVL-PV against variants trained without structured reasoning. We evaluated two training regimes: (1) frozen vision encoder regime, where only language model parameters are updated, enabling assessment of reasoning mechanisms independent of visual adaptation, and (2) unfrozen vision encoder regime, where both vision and language components are jointly optimized. In both regimes, we compared standard supervised fine-tuning (SFT) without reasoning against RSFT followed by 2PRE. 

Sequential refinement through visual adaptation and reasoning alignment yielded cumulative improvements from a zero-shot baseline of 8.1\% to a final peak of 93\%, with all results reported as the mean of five independent runs to ensure reproducibility. We observed distinct performance trajectories across the two regimes, in the (1) frozen vision encoder regime, the performance was driven primarily by 2PRE, while SFT reached 37.9\%, the addition of structured reasoning with RSFT and 2PRE nearly doubled to 70\% and with the addition of TTA, it reached 80\%, demonstrating that logical constraints can effectively compensate for static visual representations. However, the full potential of the framework was realized only when visual flexibility was combined with structured reasoning, (2) unfrozen vision encoder regime, the SFT reached 72\%, yet it was the integration of structured reasoning with RSFT that triggered a qualitative shift in performance to 91\%, with final 2PRE and TTA achieving 93\% (Fig.~\ref{fig:model_accuracy_comparison}b). This 19-percentage-point gain over SFT indicates that explicit reasoning acts as a critical regularizer, preventing the model from overfitting to superficial visual correlations. Across both training regimes, the addition of reasoning supervision consistently yields the largest marginal improvements, confirming that diagnostic reasoning, rather than model capacity alone, drives the observed performance gains. Full prediction workflow, complete hyperparameter settings and qualitative examples of reasoning progression across training stages are provided in Supplementary Note 2.

These gains extend consistently across defect categories. The per-class F1 scores (Fig.~\ref{fig:model_accuracy_comparison}c) show that REVL-PV achieves near-perfect performance across the defect categories and maintains competitive performance even for the most visually ambiguous defect types, outperforming both YOLOv11x\cite{khanam2024yolov11} and Gemini 2.5 Flash\cite{comanici2025gemini}. 

\noindent{}\textbf{REVL-PV demonstrates superior selective prediction under uncertainty.} Classification accuracy alone does not fully capture a model's suitability for deployment in safety-critical settings, where the cost of a confident wrong prediction may exceed the cost of abstaining. To evaluate this dimension, we assessed selective prediction performance using Risk-Coverage (RC) analysis, which quantifies how effectively a model's confidence scores separate correct from incorrect predictions across varying coverage thresholds. Full details regarding logit extraction and the post-hoc calibration techniques required to mitigate reinforcement learning-induced overconfidence are provided in Supplementary Note 3.

We first evaluated the models across all coverage levels, where REVL-PV achieves a lower overall Area Under the Risk-Coverage Curve (AURC) of 0.0511 compared to 0.0541 for YOLOv11x. Because low-coverage operating points are unlikely to be useful in practice, we additionally evaluated partial AURC over the 50-100\% coverage range (AURC\textsubscript{50-100}), corresponding to realistic human-AI inspection workflows (Fig.~\ref{fig:model_accuracy_comparison}d). In this practically relevant high-coverage regime, REVL-PV maintains a substantially lower partial AURC than YOLOv11x (0.0261 versus 0.0390). At deployment-relevant coverage thresholds of 50\%, 70\%, and 90\%, REVL-PV yields risks of 4.7\%, 4.6\%, and 5.7\%, respectively, compared with 7.0\%, 7.0\%, and 8.4\% for YOLOv11x. At full coverage, the gap widens further to 8.5\% versus 13.4\%. A transient increase in risk appears at low coverage (5-15\%), driven mainly by confident errors between visually similar EL defect pairs, especially \textit{Finger} and \textit{Thick Line}. Because these classes share near-identical morphological signatures, REVL-PV can assign high confidence to the wrong sub-category based on genuine but weakly discriminative visual evidence. This pattern suggests a limit of visual separability rather than spurious overconfidence. A visual comparison demonstrating the near-identical signatures of these defects is provided in Supplementary Note 3.

\noindent{}\textbf{REVL-PV maintains robust performance across corruption categories and perturbations.} To assess resilience under realistic deployment conditions, we evaluated REVL-PV against three categories of image corruptions designed to simulate field acquisition variability. These consists of additive Gaussian noise (simulating sensor noise), Gaussian blur (simulating focus errors and motion artifacts), and geometric transformations including random flips, rotations (up to 45\degree), affine shear, and occlusions (simulating viewpoint variations, panel misalignment, and partial obstruction). Each corruption type was applied at five progressive severity levels to the entire test set of images, with severity 1 representing minimal perturbation and severity 5 representing extreme degradation. Detailed quantitative robustness analyses across all three corruption types and five severity levels, including full degradation curves, are provided in Supplementary Note 4.

\begin{figure}[H]
    \phantomsection
    \captionsetup{
        justification=raggedright,
        singlelinecheck=false,
        labelfont=bf
    }

    \makebox[\textwidth][c]{%
        \includegraphics[width=1\textwidth]{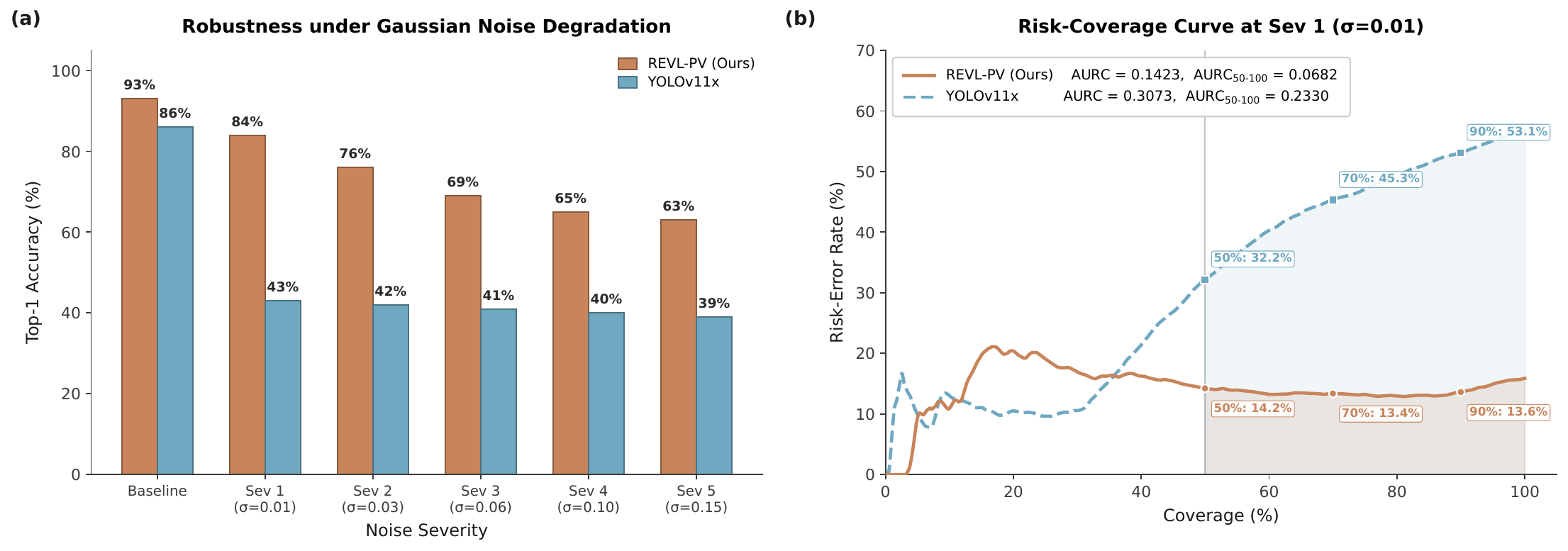}
    }

    \caption{\textbf{Model robustness and predictive reliability under physical signal degradation.}
    \textbf{a. Robustness under Noise Degradation.}  While the baseline YOLOv11x fails catastrophically at minimal noise levels (dropping from 86\% to 43\% accuracy at Severity 1, $\sigma=0.01$), the reasoning-aware REVL-PV framework degrades gracefully. Even under the most severe visual corruption (Severity 5, $\sigma=0.15$), REVL-PV retains a strong 63\% top-1 accuracy, outperforming YOLOv11x by 24 percentage points.
    \textbf{b. Risk-Coverage Curve at Noise severity 1.} In the operationally relevant high-coverage regime ($\ge$50\%), REVL-PV maintains stable uncertainty calibration, holding a consistently low risk profile (e.g., a 13.6\% error rate at 90\% coverage). In contrast, YOLOv11x exhibits severe overconfidence and escalating risk-error rates, surging to 53.1\% at 90\% coverage. This stability is quantified by the partial Area Under the Risk-Coverage curve ($\text{AURC}_{50-100}$), where REVL-PV achieves 0.0682 compared to YOLOv11x's 0.2330.
    }

    \label{fig:robustness}
\end{figure}

REVL-PV consistently outperformed YOLOv11x across all corruption settings (Fig.~\ref{fig:robustness}a). Under additive noise, both models began from clean-image accuracies of 93\% and 86\%, respectively, but diverged sharply even at the mildest perturbation. At severity 1 (variance = 0.01), YOLOv11x dropped to 43\%, whereas REVL-PV retained 84\% accuracy. This gap persisted across increasing severities, with REVL-PV remaining functional even at severity 5 (variance = 0.15), where it achieved 63\% accuracy. These results suggest that reasoning-aware supervision improves robustness to degraded inputs.

We further evaluated selective prediction under mild noise using risk-coverage analysis (Fig.~\ref{fig:robustness}b). REVL-PV achieved substantially lower AURC than YOLOv11x and maintained markedly lower risk across the deployment-relevant 50-90\% coverage range. Whereas YOLOv11x risk increased sharply from 32.2\% to 53.1\%, REVL-PV remained comparatively stable (14.2\% at 50\% coverage, 13.4\% at 70\%, and 13.6\% at 90\%). This indicates that REVL-PV preserves not only higher accuracy under corruption but also more reliable uncertainty estimates, an important property for practical photovoltaic inspection.

\begin{figure}[H]
    \phantomsection
    \captionsetup{
        justification=raggedright,
        singlelinecheck=false,
        labelfont=bf
    }

    \makebox[\textwidth][c]{
        \includegraphics[width=0.8\textwidth]{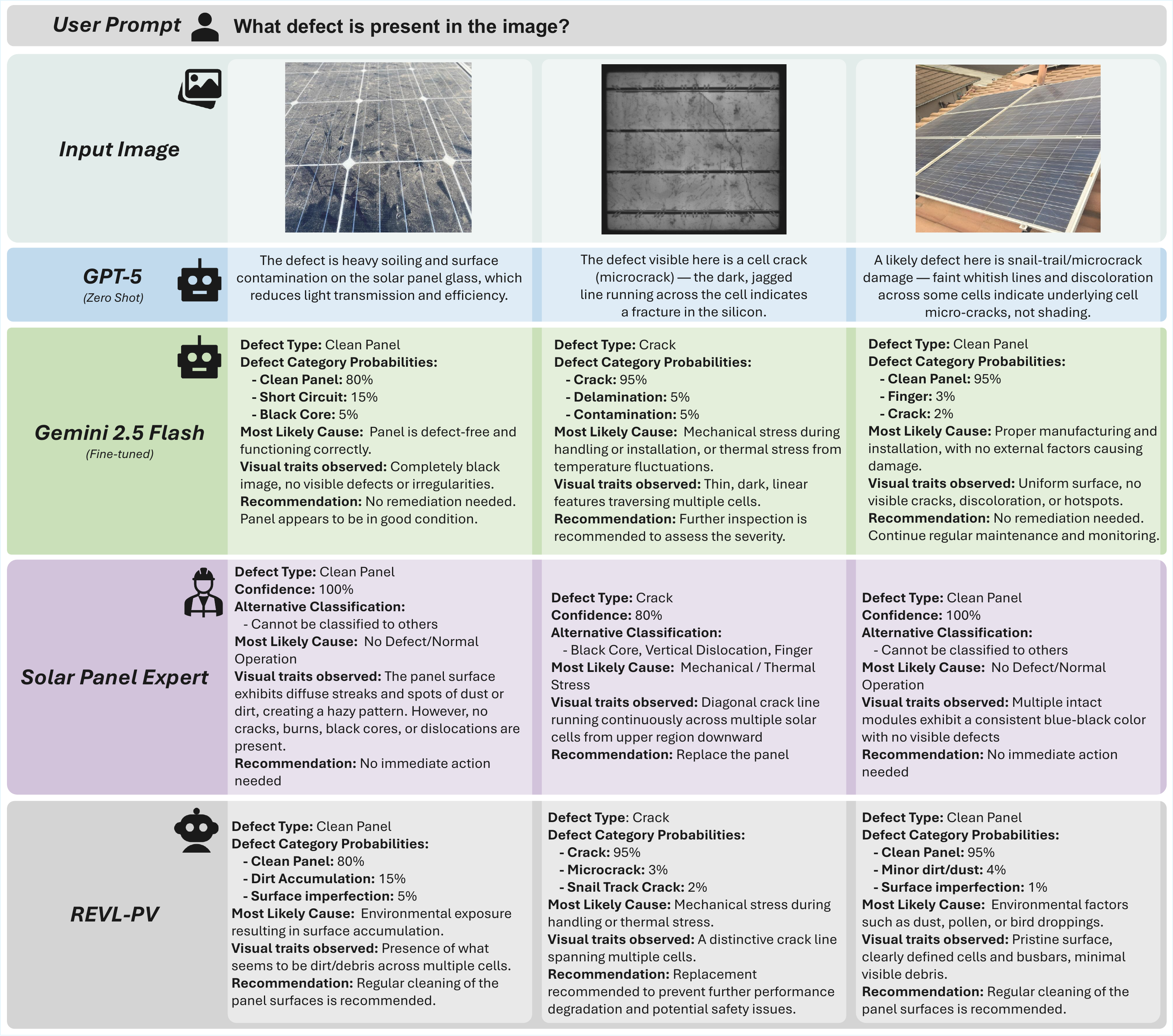}
    }

    \caption{\textbf{Qualitative comparison of diagnostic outputs.} Representative predictions from GPT-5 (zero-shot), Gemini 2.5 Flash (fine-tuned), a certified solar panel expert and REVL-PV across three test cases. General-purpose models like GPT-5 hallucinated in Case 3, misclassifying a ``Clean Panel’’ as ``snail-trail/microcrack'', a false positive. Similarly, in the case of Gemini 2.5 Flash, despite being fine-tuned exhibit critical failures including fabricated visual features and arithmetically inconsistent confidence scores. To establish a ground-truth baseline, the expert evaluated the images under fully blind conditions using structured assessment that mirrored the model's unified eight-class defect taxonomy and output format. REVL-PV produces fully structured outputs including defect type, calibrated probability distributions, root cause, and recommended action, with no hallucinations across all cases.}

    \label{fig:qualitative_comparison}
\end{figure}

\begin{figure}[H]
    \phantomsection
    \captionsetup{
        justification=raggedright,
        singlelinecheck=false,
        labelfont=bf
    }

    \makebox[\textwidth][c]{%
        \includegraphics[width=0.9\textwidth]{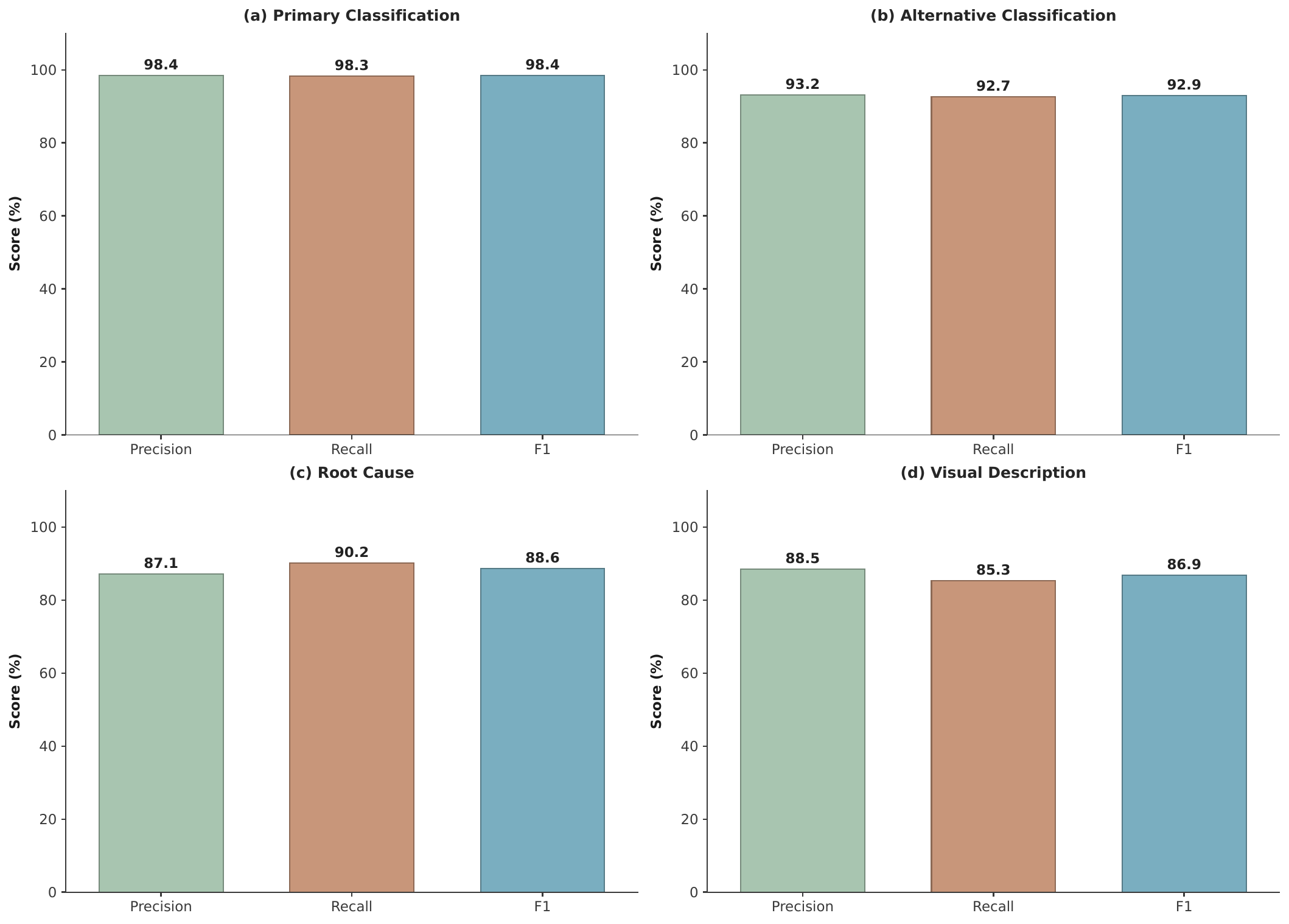}
    }

    \caption{\textbf{Semantic alignment of generated diagnostic reports.} 
    The charts display BERTScore metrics (Precision, Recall, F1) measuring the linguistic similarity between the model's textual output and the human expert's ground truth descriptions across four diagnostic components. \newline
    \textbf{a–b. Classification Alignment.} The near-perfect scores in Primary Classification (F1: 98.4\%) and Alternative Classification (F1: 92.9\%) indicate that the model utilizes the exact industry-standard terminology required for formal reporting. 
    \textbf{c. Causal Reasoning.} The high Recall (90.2\%) in the Root Cause analysis demonstrates that the model successfully retrieves the vast majority of underlying factors identified by experts (e.g., thermal stress, manufacturing defects) while ensuring no critical warnings are omitted. 
    \textbf{d. Descriptive Accuracy.} The strong performance in Visual Description (F1: 86.9\%) confirms the model's ability to articulate complex visual traits with a level of detail comparable to expert analysis.}

    \label{fig:bertscore_evaluation}
\end{figure}

\noindent{}\textbf{REVL-PV produces diagnostic reasoning consistent with expert assessment.} To validate the quality of the model's reasoning beyond classification accuracy, we conducted a blind concordance study against a certified solar panel expert, evaluating alignment across four core reasoning dimensions comprising primary classification, alternative classification, root cause, and visual description. This assessment utilized a manually curated subset of images featuring both unambiguous and visually challenging defects to ensure a rigorous evaluation of diagnostic logic. The expert was blinded to model predictions and used a unified eight-class taxonomy alongside free-text fields for root cause and visual evidence (Fig.~\ref{fig:qualitative_comparison}). Semantic similarity between model-generated and expert-authored reasoning was quantified using BERTScore\cite{zhang2020bertscore} (Fig.~\ref{fig:bertscore_evaluation}). The model achieved near-perfect agreement on primary classification (F1 = 98.4\%) and alternative classification (F1 = 92.9\%), confirming that its defect identification logic closely mirrors expert judgment. Agreement remained strong for root cause attribution (F1 = 88.6\%) and visual trait description (F1 = 86.9\%). The consistency across all four dimensions yields an average aggregate BERTScore F1 of 91\%, confirming that REVL-PV does not merely replicate classification outcomes but reproduces the underlying diagnostic process, ensuring that automated decisions remain transparent and verifiable against professional inspection standards. Full methodological details of the concordance study and BERTScore evaluation procedure are provided in Supplementary Note 5.

By anchoring its diagnostic rationales to primary visual signatures rather than inferred functional causes, the model demonstrated strict adherence to physical evidence, confirming that the system effectively identifies and describes physical anomalies while ensuring automated decisions remain transparent and verifiable.

\noindent\textbf{REVL-PV exceeds state-of-the-art VLM capabilities.} To contextualize the diagnostic quality of REVL-PV outputs, we conducted a qualitative comparison against a certified solar panel expert and two state-of-the-art VLM, GPT-5\cite{singh2025openai} (operating in a zero-shot capacity) and Gemini 2.5 Flash\cite{comanici2025gemini} (fine-tuned on our dataset). We utilized an identical single-sentence prompt across all three computational evaluators on three representative cases (Fig.~\ref{fig:qualitative_comparison}). Both state-of-the-art models exhibited significant hallucinations that undermine their utility in professional inspection. For instance, in Case 3, GPT-5\cite{singh2025openai} hallucinated a ``snail-trail/microcrack'' defect on a clean panel, a false positive that could trigger unnecessary and costly replacements. While Gemini 2.5 Flash\cite{comanici2025gemini} was fine-tuned, it still demonstrated critical reasoning failures. In Case 1, it produced a complete black-box hallucination, claiming the image was ``completely black" despite visible features, and in Case 2, it fabricated conflicting defect probabilities that summed to an impossible 105\% (95\% crack, 5\% delamination, and 5\% contamination). REVL-PV, by contrast, produced fully structured diagnostic outputs including defect type, confidence distributions, root cause, and recommended actions across all cases with no hallucinations. This comparison confirms that domain-aware, reasoning-based supervision produces diagnostic fidelity that general-purpose models (even when fine-tuned or operating at immense scale) cannot replicate.

\section*{Discussion}

In this work, we show that structured diagnostic reasoning with domain knowledge, rather than model scale alone, is a key driver of both performance and interpretability in automated photovoltaic defect inspection. By supervising the intermediate reasoning process that links visual evidence to plausible fault mechanisms before classification, REVL-PV achieves 93\% accuracy while producing diagnostic rationales that align closely with expert judgment (average BERTScore of 91\%). These findings suggest that performance and transparency need not be competing objectives, but can instead be jointly improved through reasoning-aware multimodal learning.

To establish confidence in these findings beyond single-metric performance, we validated REVL-PV across four complementary dimensions: (1) reasoning-aware ablation, (2) cross-paradigm benchmarking against object detection, vision transformer, and vision-language baselines, (3) systemic robustness testing, and (4) expert concordance analysis, each designed to interrogate a distinct property of the framework, specifically whether structured reasoning drives performance gains, whether those gains generalize across model families, and whether the diagnostic outputs are meaningful to domain experts.

Our results further indicate that the gains of REVL-PV arise primarily from supervision structure rather than visual adaptation alone. Although supervised fine-tuning established the visual foundation, the largest improvements came from reasoning-aware supervision, including RSFT, 2PRE, and TTA. This pattern shows that logical diagnostic constraints can substantially improve performance, even when visual representations are only partially adapted, while also producing outputs that follow the Evidence$\rightarrow$Cause$\rightarrow$Action structure used by human inspectors. Together, these findings support a broader design principle: trustworthy industrial AI may depend less on increasing model capacity than on aligning supervision with domain-specific reasoning practices. Risk-coverage analysis shows that reasoning-aware supervision improves not only classification accuracy but also the reliability of predictive uncertainty. After recalibration, REVL-PV achieves lower AURC than YOLOv11x across both the full coverage range and the practically relevant high-coverage regime. This advantage becomes larger under corrupted inputs, where REVL-PV maintains substantially lower risk across deployment-relevant coverage levels. These results suggest that embedding diagnostic reasoning into learning yields confidence estimates that remain more informative under both clean and degraded imaging conditions.

These findings have direct implications for photovoltaic energy infrastructure. As solar deployment continues to scale, module inspection is becoming an increasingly important challenge for both manufacturing quality control and field maintenance. Automated systems that provide interpretable diagnostic reports could reduce operations-and-maintenance costs while improving reliability across large solar fleets. By linking visual evidence to plausible fault mechanisms, reasoning-aware systems such as REVL-PV can support engineers in auditing inspection outcomes, prioritizing maintenance actions, and enabling human-AI collaborative workflows in which routine screening is automated and ambiguous cases are escalated to experts. 

The findings presented here also reveal boundaries that future work should address. The reasoning supervision pipeline currently relies on synthetically generated prompts via large language models which, while effective, may not fully capture site-specific failure modes. Also, the current framework achieves high diagnostic accuracy using visual data alone, the framework’s multimodal inputs could be expanded beyond visual data to incorporate continuous electrical measurements by integrating operational signals, such as panel voltage, current-voltage characteristics, and real-time power output, which would empower the reasoning pipeline to directly correlate physical anomalies with their corresponding electrical fault signatures. Such integration could also resolve the confident misclassifications observed between morphologically similar defect pairs such as \textit{Finger} and \textit{Thick Line}, where purely visual features are insufficient to distinguish sub-defect boundaries but distinct electrical signatures could provide the necessary discriminative evidence. More fundamentally, REVL-PV currently treats inspection as a static single-image classification task, whereas real-world degradation is inherently longitudinal. Crack propagation, delamination progression, and performance decline unfold over years of field operation. Extending the reasoning framework to longitudinal analysis would transform it from a diagnostic tool into a prognostic one, enabling remaining useful life estimation alongside defect classification.

\section*{Methods}

\noindent\textbf{Stage 1: Dataset curation.} We constructed a multimodal PV defect dataset (Fig.~\ref{fig:model_architecture} Stage 1) comprising three distinct imaging modalities: (1) EL, (2) thermal, and (3) visible-light RGB. These were sourced from the publicly available datasets namely, PVEL-AD\cite{su2022pvel}, PVMD\cite{bello2024photovoltaic}, and  Kaggle and Roboflow repositories. Most of the samples were relabeled according to a unified eight-class taxonomy: \textit{Crack}, \textit{Clean Panel}, \textit{Finger}, \textit{Black Core}, \textit{Thick Line}, \textit{Horizontal Dislocation}, \textit{Vertical Dislocation}, and \textit{Short Circuit}. Images underwent automated integrity checks to filter corrupted files. Ambiguous samples lacking clear ground-truth annotations were excluded after review. The final unified corpus comprised 12,847 images distributed across EL, thermal, and RGB modalities. The complete label standardization mapping is provided in Supplementary Note 6.

\noindent\textbf{Stage 1: Class balancing and augmentation.} Initial analysis of the pre-curation corpus revealed substantial class imbalance, driven primarily by the \textit{Crack} category, which initially comprised 4,474 images aggregated across all three modalities comprising 38.2\% of samples while \textit{Short Circuit} accounted for only 4.1\%. To mitigate this, we implemented a two-stage rebalancing protocol. We applied stratified undersampling to \textit{Crack} at a 40\% retention quota to each constituent sub-dataset (EL: $3{,}000\rightarrow1{,}200$, Thermal: $280\rightarrow112$, RGB: $1{,}194\rightarrow478$), yielding 1,790 retained images that preserved the original distribution of imaging conditions. 

All minority classes were augmented to 1,500--1,800 images using random rotation ($\theta\in\{\pm15^{\circ},\pm30^{\circ}\}$), horizontal and vertical flipping, brightness scaling ($\alpha\in\{0.8,1.2\}$), contrast enhancement ($\alpha=1.3$), and additive uniform noise (factor = 0.1) to simulate varying exposure conditions and sensor noise. 

The dataset was split into training (70\%), validation (15\%), and test (15\%) sets using stratified random sampling to preserve class and modality distributions. Test set composition was held constant across all experiments to ensure fair comparison.

\noindent\textbf{Stage 2: Structured prompt generation.} To explicitly operationalize Chain-of-Thought (CoT) reasoning, each training image was paired with a structured prompt (Fig.~\ref{fig:model_architecture} Stage 2) comprising two mandatory components: (1) \texttt{<think>} block for intermediate logical deductions, where the model must identify visual evidence and rule out false positives before committing to a decision and (2) \texttt{<answer>} block  containing the predicted defect class, confidence probabilities, root-cause analysis, a description of visual evidence, and recommended action. This design treats diagnostic rationale as a prerequisite for classification rather than a post-hoc rationalization.

 We synthesized these prompts using Google Gemini 2.5 Pro\cite{comanici2025gemini} via the Vertex AI API. For each image, the model was provided with the ground-truth label and a system instruction emphasizing conciseness and logical specificity. Generation parameters were fixed at temperature = 0.7, top-$p$ = 0.95. The full system prompt template used for generation and inference is provided in Supplementary Note 7.
 
 Quality assurance was two-fold: (1) automated regex pattern matching filtered outputs for schema compliance, followed by (2) manual expert review of 500 random samples to certify the absence of hallucinations or data leakage. Hallucination was operationally defined as any instance where the reasoning chain contained visual attributes absent from the ground-truth label, causally inconsistent explanations contradicting known photovoltaic failure mechanisms, reasoning collapse (identified as explicit acknowledgment of prompt-driven classification rather than evidence-driven inference), or systematic use of N/A placeholders substituting for required diagnostic steps.

\noindent\textbf{Stage 2: RSFT with synthetic prompts.} We utilized Qwen-2.5-VL-3B\cite{bai2025qwen25vl} as the backbone for REVL-PV. The architecture comprises a 27-layer vision encoder, a two-layer MLP projection module, and a 36-layer language model with grouped-query attention. The model was trained on image-prompt pairs via next-token prediction, maximizing the log-likelihood of the ground-truth responses. (Fig.~\ref{fig:model_architecture} Stage 2)
\[
\mathcal{L}_{\text{RSFT}} = -\sum_{t=1}^{T}\log P\!\left(\mathbf{y}_t\mid \mathbf{y}_{<t},\mathbf{x},\mathbf{I},\boldsymbol{\theta}\right),
\]
where $\mathbf{I}$ is the input image, $\mathbf{x}$ is the text prompt, $\mathbf{y}_t$ is the $t$ target token, and $\boldsymbol{\theta}$ represents model parameters. Training utilized fine-tuning scripts from LlamaFactory\cite{zheng2024llamafactory} and AdamW ($\mathrm{lr}=1\times10^{-4}$), a global batch size of 64 (per-device batch size of 2 with 2 gradient accumulation steps), and ran for 3 epochs on 16$\times$ A100 GPUs with DeepSpeed ZeRO\cite{rajbhandari2020zero} Stage 3. 

To isolate the contributions of visual adaptation versus reasoning alignment, we evaluated two distinct training regimes: (1) frozen vision encoder (optimizing only the language head) and (2) unfrozen vision encoder (enabling end-to-end spectral adaptation to PV-specific imaging modalities)

\noindent\textbf{Stage 3: 2PRE Phase 1 (Text-only reasoning).} To decouple logical structuring from visual recognition, we performed an initial stage of 2PRE on 9,500 synthetic text-only prompts (Fig.~\ref{fig:model_architecture} Stage 3). We employed the REINFORCE Leave-One-Out (RLOO) algorithm, which is more sample-efficient than standard rejection sampling for reasoning tasks. More details about synthetic text generation along with representative synthetic scenario templates is provided in Supplementary Note 7.

For each prompt $x$, the model generated $K=6$ parallel reasoning chains $\{y_1, ..., y_K\}$. These were evaluated by a rule-based verifier that assigned a binary reward $R(y_i) \in \{0, 1\}$ based on logical coherence. The model parameters $\theta$ were updated to maximize the expected reward using the RLOO advantage estimator, which uses the mean reward of peer samples as the baseline:
\[
\nabla J(\boldsymbol{\theta}) \approx \frac{1}{K} \sum_{i=1}^K \left( R(\mathbf{y}_i) - \frac{1}{K-1} \sum_{j \neq i} R(\mathbf{y}_j) \right) \nabla_{\boldsymbol{\theta}} \log \pi_{\boldsymbol{\theta}}(\mathbf{y}_i \mid \mathbf{x})
\]
Drawing architectural inspiration from the LMM-R1\cite{peng2025lmm} framework, 2PRE training was executed using OpenRLHF\cite{hu2024openrlhf} on 8$\times$ A100 GPUs (LR = $1\times10^{-6}$) DeepSpeed ZeRO\cite{rajbhandari2020zero} Stage 3. This stage effectively initialized the model's ``thinking'' patterns before the introduction of visual data.

\noindent\textbf{Stage 3: 2PRE Phase 2 (Multimodal Grounding).}
To align the model's latent reasoning with expert diagnostic protocols, we trained this checkpoint using Proximal Policy Optimization (PPO\cite{schulman2017proximal}) (Fig.~\ref{fig:model_architecture} Stage 3). Unlike the RLOO stage, which optimized reasoning traces in isolation, this phase grounded the logic in visual evidence using a learned value function.

We implemented a rule-based reward $R(\mathbf{x},\mathbf{y})$ enforcing a strict hierarchy of objectives: Accuracy\,$>$\,Reasoning\,$>$\,Calibration. The classification reward $R_\text{cls}\in\{+1.0,-1.0\}$ assigns a binary signal based on match with ground truth. Its magnitude establishes correctness as the dominant objective, preventing the model from optimizing for plausible-sounding but factually incorrect reasoning. The reasoning-structure reward:
\[
R_\text{steps} = 0.5 \times \frac{N_\text{steps}}{7}
\]
awards partial credit for sequential diagnostic markers, where $N_\text{steps}$ is the count of unique reasoning steps and the target of seven steps reflects preliminary analysis showing that valid
diagnostic chains typically resolve within 5-7 logical leaps. The weighting factor of 0.5 ensures a perfect reasoning chain is insufficient to offset a classification error. A calibration bonus
$R_\text{prob} = +0.3$ is awarded when the response includes a structured probability distribution, empirically tuned to incentivize confidence reporting without overriding the primary classification signal. A formatting penalty $R_\text{pen} = -0.5$ is applied when critical tags are missing or the reasoning chain is truncated ($N_\text{steps}<4$). The total reward
\[
R_\text{total}=\operatorname{clamp}\!\left(R_\text{cls}+R_\text{steps}+R_\text{prob}+R_\text{pen},\,-1.0,\,1.0\right)
\]
was maximized subject to a KL-divergence constraint against the
reference policy $\pi_\text{ref}$:
\[
J(\boldsymbol{\theta})=\mathbb{E}_{(\mathbf{x},\mathbf{y})\sim\pi_{\boldsymbol{\theta}}}\!\left[R_\text{total}(\mathbf{x},\mathbf{y})-\beta\log\frac{\pi_{\boldsymbol{\theta}}(\mathbf{y}\mid \mathbf{x})}{\pi_\text{ref}(\mathbf{y}\mid \mathbf{x})}\right]
\]
Training used PPO with clipped surrogate loss and Generalized Advantage Estimation (GAE\cite{schulman2016high}), with actor $\mathrm{lr}=1\times10^{-6}$, global batch size\,16, and $\beta=0.001$. To manage the memory footprint of the 3B parameter model alongside the value and reference models, we employed a heterogeneous GPU allocation on 8$\times$A100s (4 GPUs for the Actor, 2 for the Critic, and 1 each for the Reference model) and vLLM inference engine. Optimization was stabilized using DeepSpeed ZeRO\cite{rajbhandari2020zero} Stage 3 with CPU offloading for optimizer states.

\noindent\textbf{Stage 4: TTA and robust inference.} At inference (Fig.~\ref{fig:model_architecture} Stage 4), we implemented a TTA strategy to ensure spatial invariance and capture both overarching contextual features and subtle localized anomalies. Each input image was systematically processed across six distinct spatial views comprising the original full-resolution image, one central crop, and four corner crops. To synthesize these multi-view predictions, the final classification was determined by majority vote over the crop predictions. This aggregation strategy was explicitly engineered to prioritize diagnostic sensitivity to minute structural failures. Consequently, localized defects detected by a majority of the cropped views strictly superseded `clean panel' predictions derived from the full image. Furthermore, to maximize defect recall, if the full image view predicted a specific anomaly and even a single localized crop confirm this finding, the defect classification was retained, explicitly overriding a ‘clean’ consensus among the remaining crops. This tiered approach effectively minimizes false negatives critical to real-world operational deployment. Conversely, in instances where the localized cropped views failed to achieve a decisive consensus, the full-image prediction served as the definitive tiebreaker to stabilize the final diagnostic output. Following this spatial aggregation step, text generation for the accompanying reasoning chains employed nucleus sampling ($p=0.9$, temperature = 0.7) and was strictly capped at a maximum of 768 tokens to ensure concise and highly deterministic rationales. To validate the efficacy of this conflict resolution strategy, an ablation study comparing alternative TTA aggregation methods is provided in Supplementary Note 8.

\noindent\textbf{Computational infrastructure.} All training experiments were conducted on a distributed cluster of 16 NVIDIA A100 40GB GPUs. Total training time was approximately 80 GPU-hours. Inference benchmarks were performed on a single A100 GPU using Flash Attention 2.

\sloppy
\section*{Data availability} 
All datasets used in this study are publicly available. PVEL-AD is accessible at \url{https://github.com/binyisu/PVEL-AD}, PVMD at \url{https://data.mendeley.com/datasets/5ssmfpgrpc/1}, and the RGB imagery datasets are available via  Kaggle at \url{https://www.kaggle.com/datasets/hemanthsai7/solar-panel-dust-detection} and Roboflow at \url{https://universe.roboflow.com/solar-panels-rgww4/fyp1-cracked-solar-panels}.

\section*{Code availability}
The source code for our framework is available at \url{https://github.com/mistrydev6/REVL-PV}

\section*{References}
\bibliography{refs/references}

\section*{Acknowledgments}
This work was supported in part by the National Science Foundation under grants IIS-2246157, FMitF-2319243, and the Department of Energy under grant DE-CR0000042. The project was also supported by computational resources provided by the NSF ACCESS and Argonne National Lab.

\section*{Competing interests}
The authors declare that they have no competing interests.

\section*{Author contributions}
R.W. conceived and designed the project. R.W. and D.M. developed the REVL-PV framework. D.M. performed the validation and prepared the manuscript. F.Q., B.C., F.L., C.C., M.S., and R.W. edited and approved the manuscript.

\end{document}